# On the Semi-Markov Equivalence of Causal Models

**Benoit Desjardins**
University of Pittsburgh
*benoit@med.pitt.edu*

## Abstract

The variability of structure in a finite Markov equivalence class of causally sufficient models represented by directed acyclic graphs has been fully characterized. Without causal sufficiency, an infinite semi-Markov equivalence class of models has only been characterized by the fact that each model in the equivalence class entails the same marginal statistical dependencies. In this paper, we study the variability of structure of causal models within a semi-Markov equivalence class and propose a systematic approach to construct models entailing any specific marginal statistical dependencies.

## 1  Introduction

One of the most central problems in scientific research is the search for explanations to some aspect of nature. This often involves a cycle of data gathering, theorizing, and experimentation. In many scientific fields data comes in the form of statistical distribution information, representing the value of different features for a sample in a population. One of the tasks in research is to discover some structure in that data. In particular, one is interested in finding something about the causal processes explaining the statistical data, in the form of a theory or a model of the aspect of nature under study. Such causal model can then be used as a basis for explanation and experimentation.

Although traditional statistical approaches are excellent for finding statistical dependencies in a body of data, they prove inadequate at finding the causal structure in the data [8, 11]. New graphical algorithmic approaches have been proposed to automatically discover the causal structure in a body of data, given certain hypotheses [2, 5, 6, 9, 12, 14]. Graph theory [17] has now emerged as a mathematical language for causality. Graphs provide a strong notational system for concepts and relationships not easily expressed by equations and probabilities, and there has been an increased understanding of the fundamental relationships between graphs, causality, and probability.

Constraint-based approaches [9, 14] use the conditional independence relations present in a body of statistical data to graphically infer the causal structure present in the data. Based on strong connections between graph theoretic properties and statistical aspects of causal influences, fundamental assumptions about the data (the Markov condition and the faithfulness condition) are used to infer a graphical structure, which represents statistical dependency relations on a set of variables. Such (marginal) dependency graph is used to construct models describing the exact causal relations in the data. Since causal relations cannot be inferred from statistical data alone, a marginal dependency graph is entailed by a possibly infinite equivalence class of models representing competing causal alternatives.

Under the assumption of causal sufficiency, there is a one to one correspondence between statistical dependency and direct causal relation. Two causal models without latents which entail the same statistical dependencies are called (local) **Markov equivalent**. Every Markov equivalence class of models is finite, and the variability of structure of models within the same Markov equivalence class is very limited. Indeed any two Markov equivalent causal models always share the same causal connections between their variables, but the direction of the causal influences can vary [1, 7, 16].

If the data contains correlated errors, the assumption of causal sufficiency fails, and latent variables must be introduced in models to explain the causal structure in the data [13]. Two causal models with latent variables which entail the same marginal dependencies are called **semi-Markov equivalent**. However marginal statistical dependency does not entail direct causal connection. Given a model $M$ with latent variables, the set of competing semi-Markov equivalent alternatives is not only large, it is infinite. One can indeed always add new latent variables to a model without changing the pattern of statistical dependencies between the measured variables. Although there is a well defined test to determine the semi-Markov equivalence of two causal models with latent variables (they must entail the same marginal dependency graph) [14, 15], there is currently no systematic approach for exploring the variability of structure within a semi-Markov equivalence class of causal models.

In this paper, given an infinite semi-Markov equivalence class of models entailing a marginal dependency graph $G$, we present a finite subset of that class which captures the essential variability of causal structure within the equivalence class, and offer graphical rules for systematically constructing from $G$ any model in that finite subset. After demonstrating that current attempts at generating semi-Markov equivalent alternatives to any given model using local graphical transformation rules are too limited, we show that our graphical rules can also be used for that purpose.

## 2   Formal preliminaries

Causal models are best represented by graphs. We first introduce all the important graph theoretical elements required in this paper [14, 17]. We assume that the reader is already familiar with the very basic concepts from graph theory.

A **directed acyclic graph** (DAG) is a pair $(\mathbf{V}, \mathbf{E})$ such that $\mathbf{V}$ is a non empty finite set of elements called **vertices** (or variables), $\mathbf{E}$ is a finite set of ordered pairs of elements of $\mathbf{V}$ called **edges**, and in which there are no cycles. If $(V_1, V_2) \in \mathbf{E}$, then there is an edge from $V_1$ into $V_2$, represented as $V_1 \rightarrow V_2$. Figure 1 (left) shows an example of such a DAG. By convention, latent variables are represented by circles, and observables by squares. A **causal graph** is a directed acyclic graph in which for each $(V_1, V_2) \in \mathbf{E}$, $V_1$ is a direct cause of $V_2$. A **causal model** is a structure $M = <G, P>$, where $G$ is a causal graph over a set of variables $\mathbf{V}$, and $P$ is a probability distribution over $\mathbf{V}$. We often identify causal models with their causal graphs in situations not involving $P$ specifically.

The expression $I(A, \mathbf{C}, B)$ represents the conditional independence relation of variable $A$ and $B$ given the variables in set $\mathbf{C}$. A causal model $< G, P >$ satisfies the **Markov condition** if every variable in $G$ is conditionally independent of its non-parents and non-descendants given its parents. $< G, P >$ satisfies the

**faithfulness condition** if all and only the conditional independence relations true in $P$ are entailed by the Markov condition applied to $< G, P >$.

A **collider** on a path $U$ is a variable $V$ (excluding the endpoints of $U$) receiving causal edges from both its two adjacent variables on $U$ ($\rightarrow V \leftarrow$). In a causal graph, two different variables $V_1$ and $V_2$ are **d-separated** given a set of (other) variables $\mathbf{W}$ if and only if there is no undirected path $U$ between $V_1$ and $V_2$ such that every collider on $U$ has a descendent in $\mathbf{W}$ and no other variable on $U$ is in $\mathbf{W}$. They are **d-connected** if and only if they are not d-separated. An important theorem is: in a causal model $M$, if variables $V_1$ and $V_2$ are d-separated given a set of variables $\mathbf{W}$ then $I(V_1, \mathbf{W}, V_2)$ [9].

An **inducing path** (relative to a set $\mathbf{S}$ of variables) from variable $V_1$ to variable $V_2$ in a causal graph is a path $U$ from $V_1$ into $V_2$ ($V_1 \ldots \rightarrow V_2$, $V_1 \neq V_2$) such that every variable in $\mathbf{S} \setminus \{V_1, V_2\}$ is a collider on $U$, and every collider on $U$ is an ancestor of either $V_1$ or $V_2$. There is an inducing path from variable $V_1$ to variable $V_2$ if and only if $V_1$ and $V_2$ are not d-separated given any subset of $\mathbf{S} \setminus \{V_1, V_2\}$. An inducing path represents a statistical dependency implied by a causal model on a marginal distribution.

The **inducing path graph** (IPG) entailed by a causal model $M$ relative to a subset $\mathbf{S}$ of the variables in $M$ is a DAG with two kinds of edges: (1) $V_1 \xrightarrow{ip} V_2$, indicating that there is an inducing path in $M$ relative to $\mathbf{S}$ between $V_1$ and $V_2$ into $V_2$, but not into $V_1$, and (2) $V_1 \xleftrightarrow{ip} V_2$, indicating that the inducing path is also into $V_1$ (Figure 1 right). The IPG of a model $M$ represents all the dependencies implied by $M$ on a marginal distribution over a subset $\mathbf{S}$ of its variables ($\mathbf{S}$ usually corresponds to the set of observables in $M$). From each model $M$ one can compute the corresponding inducing path graph $\mathbf{IPG}(M)$, by determining the inducing paths in $M$ between its observable variables. The IPGs of any two semi-Markov equivalent causal models share the same adjacencies, but might not share the same edge orientations. The **marginal dependency graph** (MDG) entailed by a causal model $M$ (also called partially oriented IPG (POIPG) or partial ancestral graph (PAG)), is a graph which has the same adjacencies as $\mathbf{IPG}(M)$, but contains only edge orientations that are common to the IPGs of all models semi-Markov equivalent to $M$. Missing orientations are indicated by small circles.

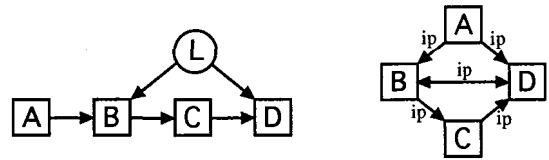

Figure 1: A DAG (left), with observables $A, B, C, D$ and latent $L$, and its entailed IPG (right).

## 3    Variability of causal models

Current constraint-based causal model discovery algorithms do not discover causal models from data. They rather discover marginal dependency graphs [14], representing an infinite equivalence class of semi-Markov equivalent causal models. We seek to study the variability of structure of models in that equivalence class, and to offer an approach to construct these models.

A marginal dependency graph $G$ represents a finite set $\mathbf{S}$ of semi-Markov equivalent IPGs entailing $G$, and contains only edge orientations that are common to all the IPGs in $\mathbf{S}$. A semi-Markov equivalence class of models can thus be finitely partitioned into infinite subclasses of models, the models within each subclass entailing the same IPG. Each IPG in $\mathbf{S}$ is a completion of $G$. Any completion of $G$ must obey certain rules to represent an IPG: it must be acyclic, and must respect certain internal consistency rules, indicated in the following theorem (proved in appendix):

**Theorem 1** *Let $G$ be an acyclic graph with $\overset{ip}{\rightarrow}$ edges and $\overset{ip}{\leftrightarrow}$ edges. Let $R_1$ be the following rule: if in $G$ there is an edge $A \overset{ip}{\rightarrow} B$ and an edge $B \overset{ip}{\leftrightarrow} C$ and $B$ is an ancestor of $C$ in $G$, then there must be an edge $A \overset{ip}{\rightarrow} C$ in $G$. $R_1$ must hold in all IPGs. Let $R_2$ be the following rule: if in $G$ there is a path of $\overset{ip}{\leftrightarrow}$ edges between variables $A$ and $B$, and every variable on that path is an ancestor of $A$ or $B$, then there must be an edge $A \overset{ip}{\leftrightarrow} B$ in $G$. $R_2$ must hold in all IPGs. An extended DAG containing $\overset{ip}{\leftrightarrow}$ edges is an IPG if and only if it is closed under the set of rules $\{R_1, R_2\}$.*

Given a marginal dependency graph $G$, the finite set of all of its acyclic completions closed under rules $R_1$ and $R_2$ is the set of all semi-Markov equivalent IPGs entailing $G$. Without causal sufficiency, any given IPG $G_i$ is entailed by an infinite set $\mathbf{S}_i$ of semi-Markov equivalent causal models. We seek to identify a finite subset of $\mathbf{S}_i$ capturing the essential variability of causal structure in $\mathbf{S}_i$ by expanding Pearl's [9] notion of "minimal models" to models with latent variables.

### 3.1    Minimal models

In [9], Pearl intuitively defines a causally sufficient minimal model as a graph in which every proper subgraph does not satisfy the Markov condition. We now extend this definition to models with latents [3].

Let $\mathbf{S}_1$ be the set of all DAGs. Let $M$ be in $\mathbf{S}_1$. Let $e$ be in $\mathbf{E}$, the set of all edges of $M$. Let $v_1, v_2$ be in $\mathbf{V}$, the set of all vertices of $M$. The operator $\mathbf{RE}_e() : \mathbf{S}_1 \rightarrow \mathbf{S}_1$ is defined such that $\mathbf{RE}_e(M)$ is the same DAG as $M$, but with edge $e$ removed. The operator $\mathbf{CV}_{v_1,v_2}() : \mathbf{S}_1 \rightarrow \mathbf{S}_1$ is defined such that $\mathbf{CV}_{v_1,v_2}(M)$ is the same DAG as $M$, but with variable $v_2$ substituted for $v_1$ in all edges of $M$, and all self-pointing edges removed. Thus $v_1$ and $v_2$ become collapsed in the new DAG. $\mathbf{RE}$ stands for "remove edge" and $\mathbf{CV}$ for "collapse variables".

Let $M$ be a model. Let $\mathbf{O}_1$ be the set of all possible operators $\mathbf{RE}_e()$, where $e$ is an edge in $M$. Let $\mathbf{O}_2$ be the set of all possible operators $\mathbf{CV}_{v_1,v_2}()$, where $v_1$ and $v_2$ are distinct variables in $M$. A **reducing transformation** $\mathbf{R}$ is a finite composition of operators in $\mathbf{O}_1 \cup \mathbf{O}_2$. If there is a reducing transformation $\mathbf{R}$ such that $\mathbf{IPG}(\mathbf{R}(M)) = \mathbf{IPG}(M)$, then the model $\mathbf{R}(M)$ is a **reduction** of $M$. The model is **minimal** (or irreducible) if for every reduction $\mathbf{R}(M)$ of $M$, we have $\mathbf{R}(M) = M$.

In other words, a minimal model is a model in which it is not possible to remove an edge or to collapse two variables in its graph without always ending up with a new model which has a different inducing path graph. Although minimal models are defined in terms of inducing path graphs for practical reasons, this turns out to be equivalent to a definition using conditional independence relations (as in the causally sufficient case) if only latent variables are allowed to be collapsed.

An edge in a causal model $M$ is an **essential edge** if it cannot be removed without changing the inducing path structure of $M$. Minimal models have only essential edges, and are the simplest models satisfying a causal theory. Minimal models do not imply simple models, but rather simplest models. They can be quite complex and contain many latent variables, including embedded and causally related latents. The essential variability of an infinite set of semi-Markov equivalent causal models is fully captured by its finite subset of minimal models [4], although a formal proof of this fact is beyond the scope of this paper.

### 3.2    Constructing causal models

We now clarify the relation between IPGs and causal models. Consider a causal model $M$ over the set $\mathbf{V}$ of variables $\{V_1, ..., V_n\}$. The IPG of $M$ over the subset $\mathbf{V}'$ of the observable variables in $\mathbf{V}$ is a graph $G$ with variables $\mathbf{V}'$. In $G$, if there is a directed edge $A \overset{ip}{\rightarrow} B$, then either $A$ is a direct cause of $B$, or there is an indirect inducing path between $A$ and $B$ in $M$. As a reminder, an inducing path between $A$ and $B$ relative to $\mathbf{V}'$ is a path $U$ from $A$ into $B$ involving a combination of variables in $\mathbf{V}'$ (observables) and variables in $\mathbf{V}'' = \mathbf{V} \setminus \mathbf{V}'$ (latents). Variables in $\mathbf{V}'$ on $U$ must all be colliders and ancestors of $A$ or $B$. Variables in $\mathbf{V}''$ on $U$ can be pretty much anything, but if they are colliders, they must be ancestors of $A$ or $B$. So the causal relations between observable variables in the causal model are strongly determined by the edges in the IPG, while the causal relations involving latent variables are not as constrained. In $G$, if there is a bidirected edge $A \overset{ip}{\leftrightarrow} B$, then by acyclicity, there must be a latent common cause T between $A$ and $B$. This might be a direct common cause, or a common cause of some other pair of variables (that could include $A$ or $B$), which spreads to $A$ and $B$ through the structure of the inducing paths.

Let $G$ be an IPG. We show that each model $M$ in the set of semi-Markov equivalent minimal models entailing $G$ is a systematic graphical transformation of $G$.

An expansion $M_i$ of IPG $G$ is a graphical transformation of $G$ made by replacing in $G$ every edge of the form $A \overset{ip}{\rightarrow} B$ by the causal edge $A \rightarrow B$, and every edge of the form $A \overset{ip}{\leftrightarrow} B$ by a latent path $A \leftarrow L_{AB} \rightarrow B$, possibly combined with a hidden edge $A \rightarrow B$ or $B \rightarrow A$ (making sure the hidden edge does not create new inducing paths or cycles). Each $M_i$ is a causal model entailing $G$ and is uniquely determined by the choice of hidden edges in $G$.

Graphical operators $\mathbf{RE}()$ and $\mathbf{CV}()$ have already been defined. We now define two additional operators. First, the operator $\mathbf{EL}_{v_1,v_2,v_3,l_1}() : \mathbf{S}_1 \rightarrow \mathbf{S}_1$ is defined such that $\mathbf{EL}_{v_1,v_2,v_3,l_1}(M)$ has the same causal graph as $M$, but if there is a pattern $[v_1 \rightarrow v_2, l_1 \rightarrow v_2, l_1 \rightarrow v_3]$ in the model where $l_1$ is a latent variable and $v_1$, $v_2$ and $v_3$ are observable variables, it is replaced by the pattern $[v_1 \rightarrow l_1, l_1 \rightarrow v_2, l_1 \rightarrow v_3, l_2 \rightarrow v_2, l_2 \rightarrow v_3]$, where $l_2$ is a new latent. Second, the operator $\mathbf{CL}_{l_1,l_2,v}() : \mathbf{S}_1 \rightarrow \mathbf{S}_1$ is defined such that $\mathbf{CL}_{l_1,l_2,v}(M)$ has the same causal graph as $M$, but if there is a pattern $[l_1 \rightarrow v, l_2 \rightarrow v]$ in the model where $l_1$ and $l_2$ are latent variables and $v$ is an observable variable, it is replaced by the pattern $[l_1 \rightarrow l_2, l_2 \rightarrow v]$. The first operator produces embedded latents ($\mathbf{EL}$ stands for "embed latent"), while the second produces causally connected latents ($\mathbf{CL}$ stands for "connect latents"). Note that the $\mathbf{EL}()$ operator introduces a new latent, because of the fact that minimal models with embedded latents can have in some rare cases more latent variables than the number of $\overset{ip}{\leftrightarrow}$ edges in the IPG [4]. In most cases in practice, a simpler version $\mathbf{EL^*}()$ is used, which does not introduce a new latent. We now have the following important theorem (the algorithm is sketched in appendix, and the full proof can be found in [4]):

**Theorem 2** *Let $G$ be an IPG. Every minimal model $M_i$ entailing $G$ is a well defined graphical transformation of an expansion of $G$ using operators* $\mathbf{RE}$, $\mathbf{CV}$, $\mathbf{EL}$ *and* $\mathbf{CL}$.

We not only have the tools to construct a minimal model entailing a given marginal dependency graph, but we have the tools to construct **all** such minimal models. The computational complexity of minimal model construction is on average $O(n^3)$ per model, where $n$ is the number of variables in the marginal dependency graph [4]. Given a marginal dependency graph $G$, the number of minimal models entailing $G$ increases exponentially with the number of edges in $G$, and strongly depends on the number of missing orientations in $G$. For marginal dependency graphs with many missing orientations, one can most efficiently constrain the set of IPGs (and therefore the number of minimal models) by using background knowledge on the observable variables.

## 4    Constructing equivalent alternatives

In the causally sufficient case, it has been formally demonstrated that any model Markov equivalent to some model $M$ can be constructed from $M$ by a sequence of edge reversal operations using only local constraints [1, 7, 16]. A natural approach with semi-Markov equivalent models is to extend this set of local graphical rules to take into account latent variables. This is what Pearl attempts in [10]. In this paper, Pearl uses $\leftrightarrow$ edges in causal graphs to indicate correlated errors between two variables. He then proposes two graphical transformation rules to construct linear semi-Markov equivalent models:

Rule 1: $X \rightarrow Y$ is interchangeable with $X \leftrightarrow Y$ if every neighbor (variable connected by a $\leftrightarrow$ edge) or parent of $X$ is adjacent to $Y$.

Rule 2: $X \rightarrow Y$ can be reversed into $X \leftarrow Y$ if every neighbor or parent of $Y$ (excluding $X$) is adjacent to $X$, and every neighbor or parent of $X$ is adjacent to $Y$.

These rules are unfortunately flawed (for both general models and linear models). Let $M$ be a causal model, in which every latent variable has no parent and is the common parent of two observable variables. In such models, subpaths $A \leftarrow L \rightarrow B$ can be replaced by edges $A \leftrightarrow B$, representing correlated errors. Pearl uses such models with $\leftrightarrow$ edges in his study of semi-Markov equivalence. We now include a counterexample to Pearl's rule 2. A counterexample to Pearl's rule 1 can be easily produced using the same principles. Consider the following causal model $M_1$ of Figure 2, where $B \leftrightarrow C$ indicates a correlated error between $B$ and $C$. In this specific case, this also corresponds to a hidden common latent cause of $B$ and $C$.

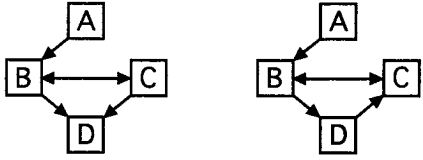

Figure 2: Model $M_1$ (left) and a (falsely) semi-Markov equivalent causal model $M_2$ (right).

According to Pearl's rule 2, the edge $C \rightarrow D$ can be reversed into $C \leftarrow D$ to produce semi-Markov equivalent model $M_2$, since the parent $B$ of $D$ is adjacent to $C$, and the neighbor $B$ of $C$ is adjacent to $D$. But $M_2$ is not semi-Markov equivalent to $M_1$. Indeed since $B$ is an ancestor of $C$ in $M_2$, there is now an inducing path from $A$ to $C$ in $M_2$, which is not present in $M_1$. Since $M_1$ and $M_2$ do not entail IPGs of same adjacencies, they cannot be semi-Markov equivalent.

Pearl's rules may be flawed, but they can be transformed into more specialized rules, which are not flawed.

Rule 1': $X \rightarrow Y$ is interchangeable with $X \leftrightarrow Y$ if every neighbor of $X$ is a neighbor of $Y$, and every parent of $X$ is a parent of $Y$.

Rule 2': $X \rightarrow Y$ can be reversed into $X \leftarrow Y$ if every neighbor of $Y$ is a child of $X$, every neighbor of $X$ is a child of $Y$, and every parent of $X$ is a parent of $Y$.

These rules are sufficient, but neither is necessary. This set can be complemented by many additional sufficient and specialized rules. Such large sets of specialized sufficient rules enable the construction of a handful of models semi-Markov equivalents to any given model. But this approach fails to consider the extensive variability of graphical structure in sets of semi-Markov equivalent causal models.

Consider the two models in Figure 3, which are semi-Markov equivalent. $M_1$ contains a latent common cause of three different observable variables, while $M_2$ contains an embedded latent. In $M_2$, observables $B$ and $D$ are neither connected directly nor through a common latent. One cannot construct $M_2$ from $M_1$ using simple edge replacement or edge reversal operations. Any single edge transformational approach to generating semi-Markov equivalent causal models is thus too limited. A larger and more powerful set of graphical operators is required. The set of operators proposed in the previous section turns out to be sufficient for that purpose. We now offer an approach to generate models semi-Markov equivalent to any model $M$ by first determining the marginal dependency graph of $M$, and then constructing minimal models entailing the same marginal dependency graph.

In [14, 15], it is formally demonstrated that two causal models with latent variables are semi-Markov equivalent if and only if they entail the same marginal dependency graph. In order to find models semi-Markov equivalent to some causal model $M$, we first extract from $M$ it's marginal dependency graph. This is a two step process: (1) determine the IPG for $M$, and (2) extract the marginal dependency graph of this IPG. The first step is a well defined graphical transformation of $M$. The second step is unfortunately an open problem, which we do not attempt to solve in this paper. Instead, we use a modified version of the TETRAD algorithm of Spirtes, et al. [14] to reconstruct it in three steps: (1) remove all edge orientations from the IPG, (2) let $A$, $B$ and $C$ be any triple of vertices such that $A$ is adjacent to $B$ and $B$ is adjacent to $C$, but $A$ is not adjacent to $C$ in the IPG; if there is no set $\mathbf{Z}$ of observable variables in $M$ such that $A$ and $C$ are d-separated by $B \cup \mathbf{Z}$, then make $B$ a collider between $A$ and $C$, otherwise make $B$ a definite non-collider between $A$ and $C$, and (3) use the TETRAD orientation rules to determine as many missing orientations as possible. The problem is open because it is not formally proven that TETRAD's set of orientation rules is complete.

Once the marginal dependency graph $G$ is determined, it suffices to use the already introduced graphical machinery to generate any causal model entailing $G$. All such models are semi-Markov equivalent to $M$.

## 4.1 Example

Figure 4 illustrates how the two semi-Markov equivalent minimal models of Figure 3 are constructed from their common marginal dependency graph (MDG): two self-consistent semi-Markov equivalent inducing path graphs are constructed ($IPG_1$ and $IPG_2$), by performing different completions of the missing orientations (small circles). The non minimal expansions $EXP_1$ and $EXP_2$ are then produced from the respective inducing path graphs by replacing correlated er-rors by common latents causes and choosing which hidden causal edges to include: no hidden causal edge for $EXP_1$ and hidden edge $C \to D$ for $EXP_2$. Note that a hidden edge $D \to C$ would not be allowed in $EXP_2$, as it would entail the inducing path $E \overset{ip}{\to} C$, which is not present in $IPG_2$. Then $M_1$ is constructed from $EXP_1$ by collapsing its three latents into a single latent $L$, and $M_2$ is constructed from $EXP_2$ by removing the non essential causal edge $B \to D$ (since the path $B \to C \leftarrow L \to D$ generates the inducing path $B \overset{ip}{\to} D$) and embedding $L$ between $A$, $C$ and $D$.

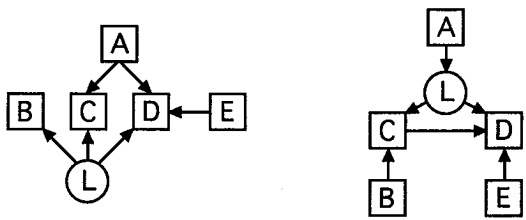

Figure 3: Two semi-Markov equivalent causal minimal models: $M_1$ (left) and $M_2$ (right).

To perform the inverse operation (dotted arrows) and construct the marginal dependency graph from each minimal model, it suffices to compute the inducing path graph for each model, and then use the Tetrad algorithm to compute their common marginal dependency graph.

The same example is used to illustrate how to construct semi-Markov equivalent alternatives to any given model: by first determining the marginal dependency graph, and then making choices of missing orientations and hidden edges to produce IPG expansions, which are then reduced by graphical transformations to produce different minimal models. There are obviously finitely many different IPG expansions produced from a marginal dependency graph. It is formally demonstrated in [4] that the total number of semi-Markov equivalent minimal models entailing a marginal dependency graph is also finite. The entire set of such minimal models can be efficiently constructed using our approach, whose main computational advantage is the important restriction of the search space of models.

## 5  Conclusion

These results are part of a larger study on the theoretical limits to reliable causal inference [4]. Constraint-based causal models discovery algorithms determine marginal dependency relations between variables, and implicitly represent sets of semi-Markov equivalent models by a single graphical structure. In this paper, we provided the formal tools to generate causal models entailing this graphical structure. Neither Markov equivalence nor semi-Markov equivalence entail distribution equivalence. By explicitly capturing the essential variability of graphical structure within a semi-Markov equivalence class of models, insights can be obtained about the formal properties of these models, and methods can be proposed to experimentally or observationally discriminate between non distribution equivalent models within the same semi-Markov equivalence class [4].

These results have also practical usefulness. Using our approach, a researcher using causal models can now, given any model satisfying a body of statistical data over a set of observable variables, automatically and systematically generate any simple alternative model explaining the data. Such alternatives might prove a valuable source of explanatory insight.

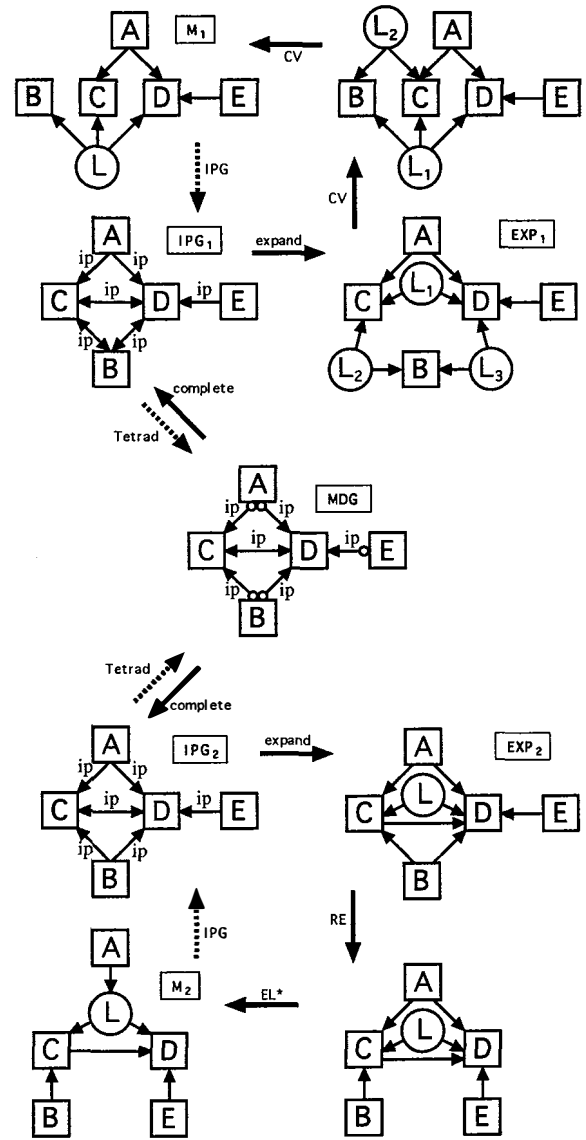

Figure 4: Construction of the two minimal models of Figure 3 from a marginal dependency graph and from each other.

# 6  Appendix: Proofs

**Theorem 1** *Let $G$ be an acyclic graph with $\xrightarrow{ip}$ edges and $\xleftrightarrow{ip}$ edges. Let $R_1$ be the following rule: if in $G$ there is an edge $A \xrightarrow{ip} B$ and an edge $B \xleftrightarrow{ip} C$ and $B$ is an ancestor of $C$ in $G$, then there must be an edge $A \xrightarrow{ip} C$ in $G$. $R_1$ must hold in all IPGs. Let $R_2$ be the following rule: if in $G$ there is a path of $\xrightarrow{ip}$ edges between variables $A$ and $B$, and every variable on that path is an ancestor of $A$ or $B$, then there must be an edge $A \xleftrightarrow{ip} B$ in $G$. $R_2$ must hold in all IPGs. An extended DAG containing $\xleftrightarrow{ip}$ edges is an IPG if and only if it is closed under the set of rules $\{R_1, R_2\}$.*

To prove the theorem, we first prove this short lemma:

**Lemma 3** *Let $M$ be a causal model with IPG $G$. If there is a directed path from variable $A$ to variable $B$ in $G$, then $A$ is an ancestor of $B$ in $M$.*

**Proof:** Let $U$ be the directed path $A \xrightarrow{ip} V_1 \xrightarrow{ip} \ldots \xrightarrow{ip} V_n \xrightarrow{ip} B$ in $G$. Each inducing path $V_k \xrightarrow{ip} V_{k+1}$ in $G$ corresponds to a path $U_k$ in $M$ that is out of $V_k$, and in which every collider is an ancestor of either $V_k$ or $V_{k+1}$. Let $W_k$ be the first collider on $U_k$ starting at $V_k$. By acyclicity $W_k$ must be an ancestor of $V_{k+1}$ and not of $V_k$. Therefore $V_k$ must be an ancestor of $V_{k+1}$ in $M$. Since this applies to every edge on $U$, then $A$ must be an ancestor of $B$ in $M$. $\quad\square$

The theorem can now be proven.

**Proof:** Assume $M$ entails IPG $G$. An edge $A \xrightarrow{ip} B$ in $G$ implies that there is an inducing path $U_1$ from $A$ into $B$ in $M$, and by the previous lemma that $A$ is an ancestor of $B$ in $M$. An edge $B \xleftrightarrow{ip} C$ in $G$ implies that there is an inducing path $U_2$ between $A$ and $B$ into both $B$ and $C$ in $M$. By concatenating the two paths $U_1$ and $U_2$ at $B$ into a new path $U$, $B$ becomes a collider on $U$, and is an ancestor of $C$. Thus $U$ is an inducing path from $A$ into $C$. So $R_1$ holds in every IPG. Let $A \xleftrightarrow{ip} V_1 \xleftrightarrow{ip} \ldots \xleftrightarrow{ip} V_n \xleftrightarrow{ip} B$ be a path between $A$ and $B$ in $G$, with each $V_k$ ancestor of $A$ or $B$ (by convention $A$ and $B$ are ancestors of themselves). The edge $V_k \xleftrightarrow{ip} V_{k+1}$ in $G$ implies that there is an inducing path $U_k$ between $V_k$ and $V_{k+1}$ into both $V_k$ and $V_{k+1}$ in $M$, with every collider being an ancestor of $V_k$ or $V_{k+1}$, and therefore of $A$ or $B$ by hypothesis. The concatenation of all these $U_k$ creates a path in which every observable is a collider, and every collider is an ancestor of either $A$ or $B$, thus an inducing path into $A$ and into $B$. So $R_2$ holds in every IPG. Closure under $R_1$ and $R_2$ is therefore necessary in an IPG. Since any observable in an inducing path (except the end-points) must be a collider, the only other combination of inducing paths not already covered that creates a new inducing path is $A \xrightarrow{ip} V_1 \xleftrightarrow{ip} \ldots \xleftrightarrow{ip} V_n \xleftrightarrow{ip} B$, with each $V_k$ ancestor of $A$ or $B$, which implies the inducing path $A \xrightarrow{ip} B$. But by acyclicity $V_1$ is necessarily ancestor of $B$, thus each $V_k$ is ancestor of $B$, which implies by $R_2$ the path $V_1 \xleftrightarrow{ip} B$. But then $R_1$ with $A \xrightarrow{ip} V_1 \xleftrightarrow{ip} B$ implies $A \xrightarrow{ip} B$. Thus closure under $R_1$ and $R_2$ is sufficient. $\quad\square$

**Theorem 2** *Let $G$ be an IPG. Every minimal model $M_i$ entailing $G$ is a well defined graphical transformation of an expansion of $G$ using operators $\mathbf{RE}$, $\mathbf{CV}$, $\mathbf{EL}$ and $\mathbf{CL}$.*

Here we only provide a sketch of the model construction algorithm. The full detailed proof, which is quite long and involves the reverse construction of minimal models of decreasing complexity, can be found in [4].

**Proof:** Given an expansion of $G$, the following sets of minimal models of increasing complexity are sequentially generated: those involving only external latents, those also involving embedded latents, and those also involving connected latents. All generated models entail the same IPG $G$.

First, given an expansion $M_1$ of $G$, all non essential edges $A \rightarrow B$ ($A$, $B$ observables) are removed from $M_1$ to produce model $M_2$ entailing the same IPG $G$. A non essential edge $A \rightarrow B$ is easily identified in $M_1$ by a simple graphical rule (lemma 4), and removing any such non essential edge from $M_1$ does not make an initially non essential edge become essential. Hidden edges are treated as special edges in the processing. The next step is to remove all non-essential latents $L$ from $M_2$ (equivalent to removing two non-essential edges per latent) to produce model $M_3$ entailing the same IPG. All non essential latents in $M_2$ are identified using simple graphical rules. Unfortunately removing a non-essential latent $L$ can occasionally make some other non-essential latent $L'$ become essential. The order of removal of non-essential latents is therefore important, and $M_3$ is usually not unique, but rather represents a very small set of alternatives.

Next, for each model $M_3$, collapsibility of latents is assessed using simple graphical rules, and latents are collapsed to produce minimal model $M_4$. Unfortunately, collapsing two latents can make two initially collapsible latents become non-collapsible. The order of collapsibility is therefore important, and $M_4$ is again not unique, but represents a finite set of alternatives. The entire set of minimal models with external latents is generated using the previous few steps.

Next, for each model $M_4$, embeddability of latents is assessed using simple graphical rules, and latents are embedded to produce minimal model $M_5$. Many combination of embeddings can be performed, with the result still being a minimal model. Thus $M_5$ is not unique. This generates the entire set of minimal models with embedded latents entailing the same IPG.

Finally, for each model $M_5$, connectibility of latents is assessed using simple graphical rules, and latents are connected to produce minimal model $M_6$. Many combination of latent connections can be performed, with the net result still being a minimal model. Thus $M_6$ is not unique. But this generates the entire set of minimal models entailing the same IPG. □

In the algorithm, simple graphical rules are used to assess features of variables or edges. Most of the power of the algorithm comes from the simplicity of such rules. Here as an example is the rule for determining the nonessentiality of edges in $M_1$. A complete description of the other rules can be found in [4]

**Lemma 4** Let $E = A \to B$ be a non hidden edge in the expansion $M_1$ of IPG $G$. $E$ is non essential (in all models of $G$) if and only if there is a path $U_1 = A \to C \leftarrow L \to B$ in $M_1$ with $C$ ancestor of $B$.

**Proof:** Assume there is such a path $U_1$ in $G_1$. $U_1$ is an inducing path from $A$ into $B$ in $M_1$ not involving $E$, so $E$ is non essential. For the converse, assume that $E$ is non essential. Then there must be an inducing path $U$ between $A$ and $B$ into $B$ in $M_1$. Then $U$ starts with $A \to C$ for some observable $C$ and continues with alternating latent common causes and observable colliders. Since by acyclicity $C$ cannot be an ancestor of $A$, it is an ancestor of $B$. Every collider on $U$ is an ancestor of $A$ or $B$, thus every collider is an ancestor of $C$ or $B$ and therefore of $B$. This implies the existence of an inducing path between $C$ and $B$ that is into both $C$ and $B$. Thus there is a $C \overset{in}{\leftrightarrow} B$ edge in $G$ and therefore there is an $A \to C \leftarrow L \to B$ path in $M_1$, with $C$ ancestor of $B$. □